\RequirePackage{fix-cm}
\documentclass[smallextended]{svjour3}       
\smartqed  
\usepackage{graphicx}

\usepackage[symbol]{footmisc}
\usepackage{cite}
\usepackage{amsmath,amssymb,amsfonts}
\usepackage{algorithmic}
\usepackage{graphicx}
\usepackage{textcomp}
\usepackage{framed,multirow}
\usepackage{mwe}
\usepackage{float}
\usepackage{comment}
\usepackage{xcolor}
\usepackage{multirow}
\usepackage{amsmath}
\usepackage{url}
\usepackage{cuted}
\usepackage{hyperref}
\usepackage{adjustbox}
\usepackage{multirow}
\usepackage{caption}
\usepackage{breqn}
\usepackage{enumitem}
\usepackage{lipsum,multicol}
\usepackage{mathtools}
\usepackage{subcaption}
\usepackage{graphicx}
\usepackage{wrapfig}
\usepackage{lscape}
\usepackage{rotating}
\usepackage{multirow}
\usepackage{tabularx}
\usepackage{appendix,float}

\definecolor{newcolor}{rgb}{.8,.349,.1}

\usepackage{longtable}

\begin{document}

\title{InstaIndoor and Multi-modal Deep Learning for Indoor Scene Recognition}

\titlerunning{Multi-modal Deep Learning for Indoor Scene Recognition in Videos}

\author{Andreea Glavan         \and
        Estefan\'ia Talavera
}

\institute{Andreea Glavan \at
              Department of Computer Science, University of Groningen \\
              \email{a.i.glavan@student.rug.nl}\\
              Corresponding author
           \and
Estefan\'ia Talavera \at Department of Computer Science, University of Groningen. Nijenborgh 9, 9747 AG Groningen, Netherlands. \\
\email{e.talavera.martinez@rug.nl}
}

\date{Received: date / Accepted: date}

\maketitle

\begin{abstract}
Indoor scene recognition is a growing field with great potential for behaviour understanding, robot localization, and elderly monitoring, among others. In this study, we approach the task of scene recognition from a novel standpoint, using multi-modal learning and video data gathered from social media. The accessibility and variety of social media videos can provide realistic data for modern scene recognition techniques and applications. We propose a model based on fusion of transcribed speech to text and visual features, which is used for classification on a novel dataset of social media videos of indoor scenes named InstaIndoor. Our model achieves up to 70\% accuracy and 0.7 F1-Score. Furthermore, we highlight the potential of our approach by benchmarking on a YouTube-8M subset of indoor scenes as well, where it achieves 74\% accuracy and 0.74 F1-Score. We hope the contributions of this work pave the way to novel research in the challenging field of indoor scene recognition.

\end{abstract}

\section{Introduction}

\textit{Scene classification} represents a growing field of research in computer vision \cite{lowry2015visual, zhou2014learning, diwadkar1997viewpoint, cheng2017remote}. It presents a challenging task due to the complexity of the appearance of locations, as well as insufficient annotated data. Despite this, there have been many works in the field of scene classification in the past years, tackling the problem by using single images \cite{espinace2010indoor, khan2016discriminative} or videos \cite{tsai2011real, silberman2011indoor}. However, there has not been much focus on the improvement of recognition of indoor scenes, which tend to share similar traits. Given this high similarity, it is difficult to automatically identify indoor scenes.

When focusing on indoor scene recognition, many works in the field are linked to robotics navigation \cite{desouza2002vision, shah1997mobile, zheng1992panoramic} and localization  \cite{liu2017scene, taira2018inloc}. Nowadays, a large amount of data is produced on a daily basis through social media \cite{gwi}. When addressing this data, the focus has been mainly on predicting the popularity which the content produced could reach \cite{li2013popularity, gelli2015image} or on predicting the credibility of the respective post \cite{singh2021predicting}. However, being able to recognize the type of location that a video depicts can support later decisions about the content or even be used in the field of forensics, specifically for extremist video detection \cite{sureka2010mining} or identification of links to drug dealing or use \cite{yang2017tracking, hassanpour2019identifying}.

Multi-modal learning aims to improve machine learning models performance by consolidating data obtained from a variety of sources into a single system. As humans, we combine information to make decisions as soon as we gather it. Following the way our brain works, our hypothesis is that by combining different modalities in an early stage of the learning model, we can later achieve a higher performance. However, multi-modal learning has different ways of accounting for the different modalities - all of them worth exploring. Examples of data modalities are visual \cite{zhang2018visual, ye2015evaluating}, audio \cite{liu1998audio, roach2001classification}, or extracted semantics in the form of detected concepts \cite{modiri2014video}. Within the context of this study, we aim to extend the body of work in the field of indoor scene classification by experimenting with multi-modal deep learning combining visual and audio information extracted from data collected from social media. 

With this in mind, we introduce a novel video dataset collected from social media, specifically Instagram. The recent surge in social media, with 90\% of young adults in the United States actively using at least one form of social media \cite{perrin2015social}, has led to an unprecedented amount of information. To the best of our knowledge, there have been no studies of indoor scene classification in the context of social media acquired information, despite the availability and realistic use cases. By collecting a dataset of indoor scene videos from social media and utilising it to train multi-modal networks, we believe a powerful and efficient means of learning can be achieved.

The lack of labelled videos for scene recognition poses a hindrance to the advancement of the field. The contributions of this work address this deficit as follows:
\begin{enumerate}
    \item We introduce and make publicly available the InstaIndoor dataset which consists of a total of 3,788 videos describing 9 indoor scenes \footnote{\label{note1} Both datasets and corresponding pipeline code are available at \url{https://github.com/andreea-glavan/multimodal-audiovisual-scene-recognition}}.
    \item We assemble a selection of YouTube videos from the YouTube-8M dataset that describe 9 indoor scenes in 900 videos.    
    \item We perform an ablation study that evaluates end-to-end multi-modal architectures for indoor scene classification.
\end{enumerate}

The rest of this paper is organized as follows: In Section \ref{sec:related_works} we present relevant studies to the field of indoor scene recognition. In Section \ref{sec:methodology} we explain our proposed methodology. Section \ref{sec:experimental} illustrates the dataset, experimental setup, and evaluation metrics, which are later presented and analysed in Section \ref{sec:results}. Finally, Section \ref{sec:conclusion} provides an overview of our work and future steps of this approach.

\section{Related Works} \label{sec:related_works}

In this section, we describe the state-of-the-art in scene recognition focusing on indoor scene identification from videos. Moreover, we give an overview of the available datasets related to this task.

Scene classification from videos started with works that vary from applying Hidden Markov Models joint with segmentation approaches \cite{huang2005joint} to segmentation techniques based on depth and intensity  \cite{silberman2011indoor}, or to Bayesian probabilistic approaches based on motion features \cite{tsai2011real}. 

The field of multi-modal learning presents itself as adequate when addressing the task of videos analysis, which can be easily described by several data modalities. Multi-modal learning has been highly successful in the medical field, both for medical imaging \cite{tan2020multimodal} and for decision support systems \cite{huang2010multimodal, harouni2018universal}. Furthermore, sensor based applications as in \cite{liu2019ensemble} benefit from fusion techniques, achieving over 77\% accuracy in machinery fault diagnosis.

The work in \cite{ngiam2011multimodal} proposed applying such a learning approach in order to maximize the information gain. By using both audio and video data extracted from the original data source in the task of video classification, the authors show that superior features for one of the modalities can be learned in the presence of multiple modalities, which leads to significant performance improvement. In this work, the authors address the task of speech processing, also known as audio-visual speech classification. The aim is to identify letters on the CUAVE \cite{patterson2002cuave} and AVLetters \cite{matthews2002extraction} datasets based on speech and visual information. Their proposed model consists of an autoencoder adapted from multitask learning applications. The model is trained both on clean and noisy data, thus becoming capable of handling noise or absence of useful features corresponding to one of the two modalities. This model achieves high accuracy with both clean and noisy audio, as well as when trained only with audio data and tested only on video data, and vice versa. This model is relevant to our research with respect to applications and similar feature usage. However, such an approach is not well suited to our purpose due to the increased complexity of scene and social videos, as opposed to their addressed speech processing at letter level.

Place recognition has been recently addressed from a visual perspective in a multitude of ways. As such, visual data can be represented and interpreted using a variety of methods \cite{leyva2021generalized,martinez2019hierarchical}. Reference \cite{leyva2021generalized} proposes recognition based on continuous similarity between image pairs via siamese CNN networks. Other approaches focus on training networks such that they learn local features. Such networks include Superpoint \cite{detone2018superpoint}, which is based on a one encoder and two decoder architecture for local features, and D2-Net \cite{dusmanu2019d2}, which provides both feature detection and pixel level descriptions. Another novel work proposes the NetVLAD architecture \cite{arandjelovic2016netvlad}. This architecture represents the state-of-the-art on many place recognition datasets, using a VGG16 core and a trainable differentiable VLAD layer focused on aggregation of local descriptors. Many variations have been built on top of NetVLAD, such as ESA-VLAD \cite{xu2021esa}, spatial pyramid-enhanced variations \cite{yu2019spatial}, and more. A notable network built on top of NetVLAD is the Patch-NetVLAD \cite{hausler2021patch}. This system focuses on both local and global feature descriptor extraction by employing multi-scale fusion of local features at image patch level. Unlike its predecessor, this network considers the global descriptors of the entire feature space, and it is viewpoint invariant as it accounts for spatial scoring of patch features. Given our use of image sequences, comparing every pair of sequences or extracting both local and global features can become very computationally expensive, thus we are not able to use these networks. Instead, we rely on environment descriptors focused on objects and places.

Large scale datasets are essential to the training process of deep learning models, and consequently for the identification of different scenes. There are many extensive image datasets available. To name a few, ImageNet \cite{deng2009imagenet} with over 1000 classes of both objects and places; SUN dataset \cite{xiao2010sun} with 899 different scene classes adding up to a total of 130,519 images; MSLS dataset \cite{warburg2020mapillary}, the largest outdoor focused dataset; Places365 dataset \cite{zhou2017places}, which contains 365 scene categories, covering a variety of areas indoor, outdoor, and urban; the MIT-67 Indoor Scene Recognition Dataset\cite{quattoni2009recognizing}, a 67 class dataset with 15620 total images describing both public (stores, leisure, public spaces) and private (home, working space) indoor scenes. Study \cite{toft2020long} further extends some of the aforementioned datasets and shows that hierarchical approaches and image-level descriptors can greatly increase the success rate.


Videos are a powerful media for information capture, both from a visual as well as an audio standpoint. Good quality videos can lead to more robust, overall better models. Video-based datasets tend to be used in computer vision for the recognition and prediction of activities \cite{oh2015action, lan2014hierarchical}, transcription on sound  \cite{Alayrac16unsupervised}, or question-answering \cite{MovieQA}. Among publicly available video datasets there is the HollyWood2 dataset \cite{marszalek09}, with 3669 videos, annotated for actions and scenes, with 12 and 10 classes respectively. The scene annotations contain mostly indoor spaces, thus providing a broad benchmark in realistic settings. Other datasets, such as Epic Kitchens \cite{damen2018scaling} and YouCook2 \cite{zhou2018towards}, provide a multitude of action oriented videos, specifically focusing on cooking techniques as well as kitchen variations. 

Several of the available video datasets provide action or object centric annotations, but not necessarily scene focused. Other notable video datasets are: the Youtube-8M dataset \cite{abu2016youtube}, containing over 1000 classes; the VLOG dataset \cite{Fouhey18}, collected by implicitly searching YouTube videos and cropping based on the action of interest; the MovieNet dataset \cite{huang2018trailers}, containing over one thousand annotated films; and the How2 dataset \cite{sanabria18how2}, a dataset with 80,000 instructional videos and corresponding captions and summaries. While there is a variety of video-based datasets available, the majority cater to action recognition and prediction, resulting in a lack of scene focused video datasets in the field of computer vision. Furthermore, indoor scenes tend to be overlooked in favour of outdoor scenes due to the applications in localization and robot navigation.

Despite the above described models, there is a distinct lack of research in the field of indoor scene recognition with respect to videos. Our goal is to extend the body of work in this field by relying on multi-modal learning techniques. We explore this by analysing available videos gathered from social media.

\section{Multi-modal Learning for Video Classification}
\label{sec:methodology}

In this section, we present our multi-modal learning methodology concerning the use of two distinct modalities, in the form of visual video frames and transcribed speech as text data. 

\subsection{Visual Modality}

\textit{Visual data}, consisting of frames which are captured from the original input video at established intervals of time, serves as an immediate choice in terms of modalities due to the information-preserving nature of images. Images capture relevant semantic information with respect to scenes due to their ability to encompass features such as objects, shapes, colors, and more. Looking at videos as a continuous collection of images captured at a high frame rate, similarity between frames is indirectly proportional to the time between them. Thus, we sub-sample the videos at a rate of one frame per second in order to preserve important information while discarding near duplicate images.
    
The sequences of extracted frames can be analysed as is using ConvLSTMs models \cite{xingjian2015convolutional}. Otherwise, these sequences can be translated into global feature vectors consisting of scene descriptors. We aim to experiment with these global scene descriptors while employing the use of pre-trained Convolutional Neural Networks (CNNs). Our approach uses the last fully-connected layer or softmax layer of the pre-trained CNNs as descriptors, which have been shown to perform exceptionally well in image classification in recent years \cite{lu2007survey}. In order to convert the frame sequence to CNN descriptors, the following process is undertaken: at video level, we input each extracted frame iteratively to the pre-trained CNN and store the resulting softmax layer values. The set of extracted probability vectors per video is then processed into a single vector by performing an element-wise sum operation on top of all the previously obtained vectors in the set. As a result, we have a 1D vector that describes the global visual context of the video.

\subsection{Text Modality} 

\textit{Transcribed textual information}, obtained from the input video's sound, can provide meaningful contextual clues in the form of conversations or powerful explanations provided by the user.
In the latter case, it is very likely that the speaker would be using either specialized terms with respect to a certain field or relevant words to the space they are in. It is possible to associate such relevant terms to their respective indoor scenes, thus allowing for better classification performance through the use of natural language processing techniques.

\begin{figure}[h!]
\centering
\includegraphics[width=1\columnwidth]{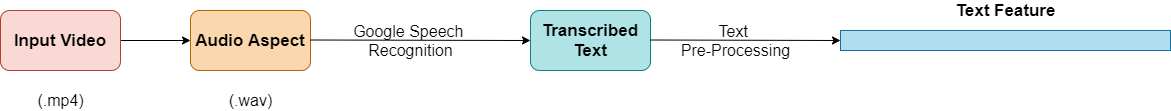}
\caption{Workflow for textual feature extraction from input video.}
\label{fig:video_to_text}
\end{figure}

The text transcription process, illustrated in Figure \ref{fig:video_to_text}, consists of the following steps: the input video is first segmented into an audio only file, which can then be transcribed into text by using existing tools, specifically the Google Speech Recognition toolkit \cite{googlest}. The Google toolkit provides the option to transcribe audio based on a user specified language, in our case English. The English language was selected for this task in order to ensure consistency within the dataset of textual features.
Once the transcribed text is obtained, the text is preprocessed by means of natural language processing techniques: it is normalized, in terms of converting all letters to lowercase, removing punctuation elements, and removing stop words. Stop words are extremely common words which hold little to no semantic value. The preprocessed text is used as a descriptive feature, which needs to be converted into a numerical format or embedding before being input to the multi-modal network. This process, also known as vectorization, is further discussed in Section \ref{sec:experimental}.

\subsection{Multi-modal learning architecture} 

\begin{figure}[ht!]
\centering
\includegraphics[width=\columnwidth]{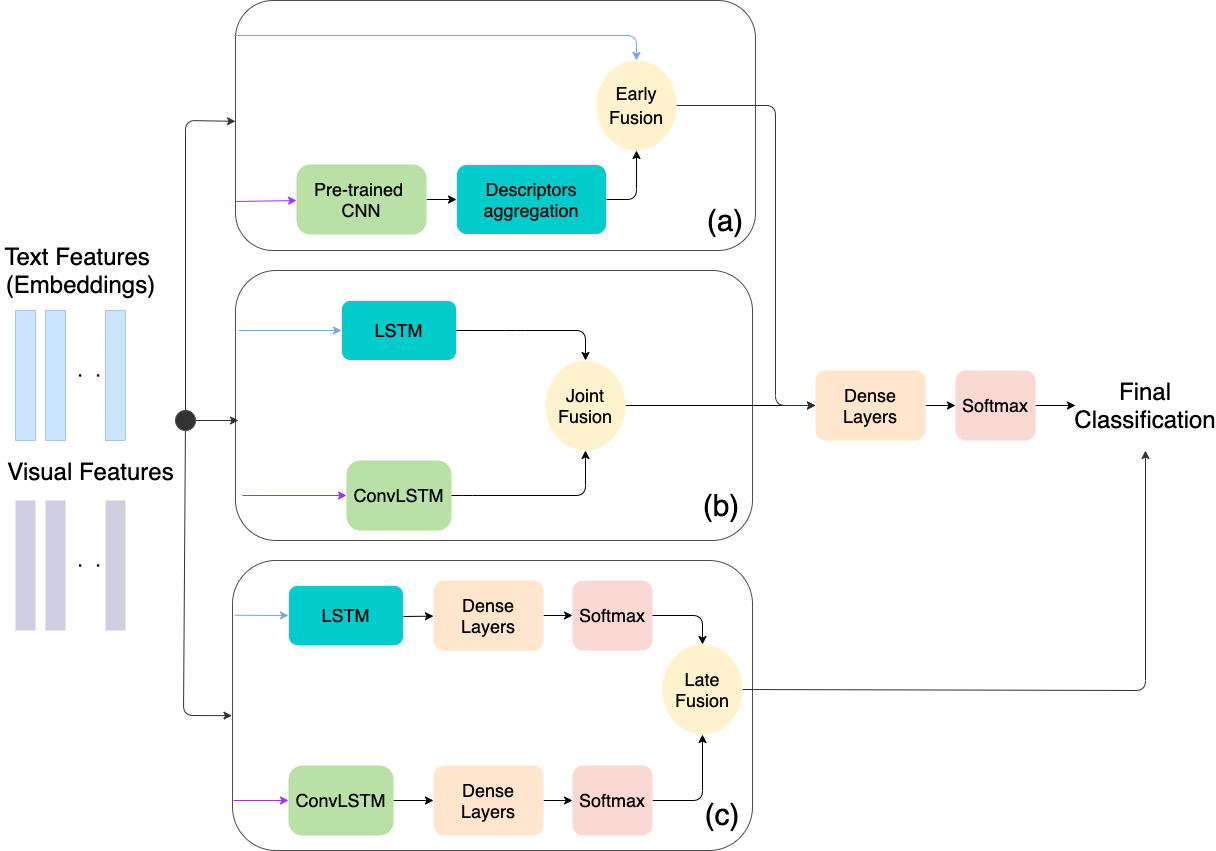}
\caption{Network architecture schemas based on fusion strategy employed. (a) Early Fusion, (b) Joint Fusion, and (c) Late Fusion.}
\label{fig:arch}
\end{figure}

Here, we describe our proposed ablation study for the evaluation of scene recognition with multi-modal deep neural network architectures.
Fusion strategies \cite{castro2020multimodal} can greatly affect the model output depending on both moment of fusion as well as type. In this work, we explore multiple fusion strategies, specifically early, joint, and late fusion, illustrated in Figure \ref{fig:arch}.

\begin{enumerate}[label=(\alph*)]

\item \textit{Early} Fusion refers to the concatenation of the input features prior to the classification process, thus the initial features per modality must be of the same dimension. Due to this constraint, frames cannot be used as features, nor can they be used after ConvLSTM processing, due to the 3 dimensional output. We rely on the extracted CNN global descriptors presented in the previous section, which are concatenated with the text features. As defined in Equation \ref{eq:early_fusion}, the modalities are concatenated element-wise. On one side, CNN features as global visual features. On the other side, we use text embeddings extracted as described above.

\begin{equation}
    EarlyFusion([t_0, t_1, ..., t_n], [v_0, v_1, ..., v_n]) = [t_0 v_0, t_1 v_1, ..., t_n v_n],
    \label{eq:early_fusion}
\end{equation}

where $t_i$ refers to the i-th text feature and $v_i$ refers to the i-th visual features.\\

\item \textit{Joint} Fusion, sometimes referred to as Hybrid or Early fusion, refers to a model where classification is performed on top of the concatenated global features of each modality. 

Unlike our previously defined early fusion, where the initial step is feature concatenation,
in this case, the features corresponding to each modality are partially classified using suitable models. In our proposed analysis we rely on LSTM and ConvLSTM with 512 units each, for text embedding and visual features analysis. From both, a global descriptor of size 512 is extracted to represent each modality. Unlike in traditional networks, a final softmax layer is not added after the dense layers. Instead, the global feature vectors of length 512 are used as is. 

The global features, as defined in Equation \ref{eq:joint_fusion}, are then concatenated and used as input to the final dense block, consisting of two layers with 256 and 128 units respectively, followed by a softmax layer of size 9 which relates to the number of output classes of our dataset.

\begin{equation}
    JointFusion([t_0, t_1, ..., t_n], [v_0, v_1, ..., v_n]) = [g_0^{t} g_0^{v}, g_1^{t} g_1^{v}, ..., g_{512}^{t} g_{512}^{v}],
    \label{eq:joint_fusion}
\end{equation}

where $t_i$ refers to the i-th text feature, $v_i$ refers to the i-th visual features, $g_i^{t}$ refers to the value at the i-th position of the global descriptor vector obtained from the textual modality classification, and  $g_i^{v}$ refers to the value at the i-th position of the global descriptor vector obtained from the visual modality classification.\\

\item \textit{Late} Fusion consists of the aggregation of the classification results at the level of each modality, meaning the features are classified separately with suitable models, LSTM or ConvLSTM each of them with 512 units. Later, the pipeline is followed by a dense fully connected layer with 256 units and a softmax layer with 9 classes. The resulting probability vectors of each model are fused according to Equation \ref{eq:late_fusion}. The final prediction is based on the obtained aggregated feature vector, with the index of the maximum value indicating the winning class.

\begin{equation}
    LateFusion([t_0, t_1, ..., t_n], [v_0, v_1, ..., v_n]) = [p_0^{t} + p_0^{v}, p_1^{t} + p_1^{v}, ..., p_n^{t} + p_n^{v}],
    \label{eq:late_fusion}
\end{equation}

where $t_i$ refers to the i-th text feature, $v_i$ refers to the i-th visual features, $p_i^{t}$ refers to the i-th value of the probability vector obtained from the textual modality classification softmax, and $p_i^{v}$ refers to the i-th value of the probability vector obtained from the visual modality classification.

\end{enumerate}

As we can observe in Fig. \ref{fig:arch}, for the implementation of (a) and (b), and in contrast to (c), the feature vector resulting from the fusion operation is further processed by means of 2 fully connected layers of size 256 and 128, respectively, followed by a softmax layer with 9 classes that indicates the final classification.

The networks share certain processing elements; Long Short Term Memory networks (LSTMs)\cite{lstm1} which provide persistent memory that can overcome time lag memory loss by gated input and output; as well as Convolutional Long Short Term Memory networks \cite{xingjian2015convolutional, si2019attention} (ConvLSTMs) which provide a similar approach to LSTMs but use internal convolutional operations as opposed to matrix multiplication operations, making them suitable for image use.

\section{Experimental framework} \label{sec:experimental}

In this section, we present the dataset and conducted experiments based on our proposed methodology. We also discuss evaluation metrics, as well as fusion strategies.

\subsection{Datasets}

\subsubsection{InstaIndoor: A novel video dataset for indoor scene recognition} 

Here, we introduce our novel InstaIndoor dataset, as one of our contributions to aiming to enrich the field of scene classification. InstaIndoor is composed of 9 indoor scene classes with 3,788 total videos collected from Instagram, distributed as according to Table \ref{tab:insta_dist}. This dataset was collected by means of web scrapping based on user-provided hashtags that describe the video content. The input hashtags corresponding to the collected data consisted of the class names or specific room names as well as possible relevant objects or activities occurring in the specific room. 

\vspace{.5in}
\begin{table}[h!]
\begin{center}
\begin{tabular}{c|cccccccccc}
Class  & \begin{rotate}{45} Cafe \end{rotate} & \begin{rotate}{45} Bar \end{rotate} &  \begin{rotate}{45}Reading Room \end{rotate}&  \begin{rotate}{45}Stadium \end{rotate}&  \begin{rotate}{45}Arcade\end{rotate} &  \begin{rotate}{45}Library \end{rotate}&  \begin{rotate}{45}Closet\end{rotate} &  \begin{rotate}{45}Beauty Salon\end{rotate} &  \begin{rotate}{45}Aquarium\end{rotate} & \begin{rotate}{45}Total\end{rotate} \\ \hline  \hline
Train & 381 & 349 & 311 & 412 & 363 &  337 & 258 & 219 & 400   &3030\\
Test & 129 & 98 & 95 & 83 & 82 & 79 & 77 & 59 & 56 & 758 \\  \hline
Total & 510  & 447 & 406          & 495     & 445    & 416     & 335    & 278          & 456 & 3788     \\

\end{tabular}
\end{center}
\caption{Entry distribution per class for the InstaIndoor dataset.}
\label{tab:insta_dist}
\end{table}

The originally gathered videos were manually filtered, with many being removed in order to maintain a high quality and relevant dataset. Videos containing content irrelevant to our search of indoor scenes, e.g., advertisements or poor quality videos, as well as those showing inconsistencies in human annotations were excluded. What is more, filtering was used to ensure that the majority of the collected videos in which there is speech involved contained mostly English speech. We also excluded videos containing only background noise or music.

\begin{figure}[h!]
\centering
\includegraphics[width=0.8\columnwidth]{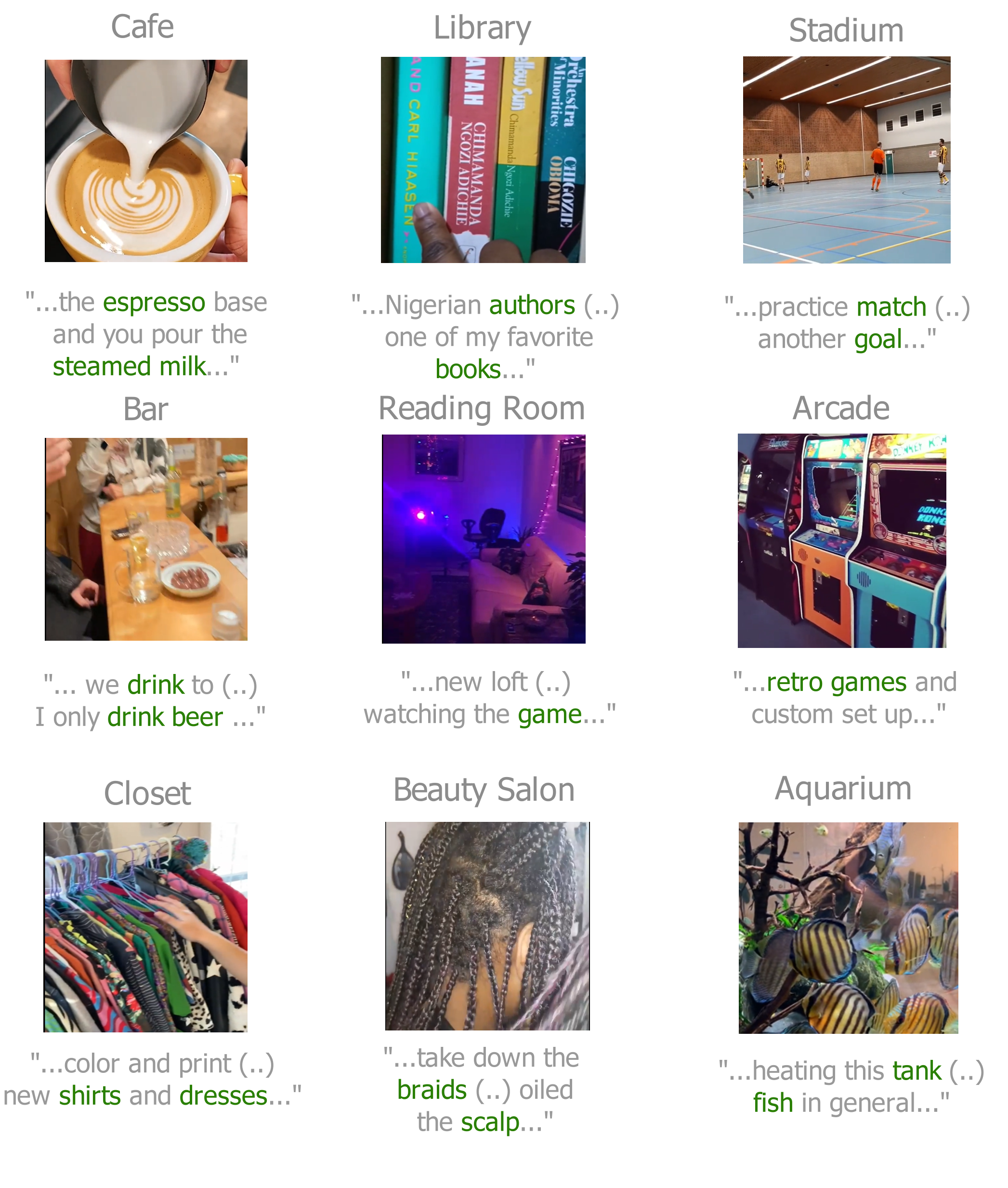}
\caption{Video samples for the classes in InstaIndoor dataset. For each class, a relevant frame of the video was selected together with the transcribed text, in which key words are highlighted.}
\label{fig:insta_img}
\end{figure}

The dataset is split according to a 7:3 ratio with respect to train:test sets. In order to evaluate the models, we used K-fold cross validation with a K value of 3. Thus, the original train set was split into a training and a validation set, corresponding to 80\% and 20\% of the original train set. This split was performed 3 times, such that random distributions were obtained over the different sets. The models were trained and validated on the new train and validation sets, and evaluated on the original test set. We compute and report the average and standard deviation of the metric presented in sub-section ~\ref{subsection:eval}.

\subsubsection{YouTubeIndoor: A YouTube-8M subset focused on indoor scenes}

We also benchmark our multi-modal classification ablation study on a subset of the Youtube-8M dataset \cite{abu2016youtube}, which is one of the largest existing publicly available video datasets. The full dataset, consisting of 1000 classes and over 6 million videos, covers classes of a variety of types, ranging from locations to actions and specific video games or objects. This comprehensive dataset has been evaluated in a variety of classification approaches, ranging from video level temporal modeling \cite{lee20182nd} to joint audio-visual mixture of experts models \cite{abu2016youtube}.

\begin{figure}
\centering
\begin{subfigure}[b]{0.47\textwidth}
   \includegraphics[width=1\linewidth]{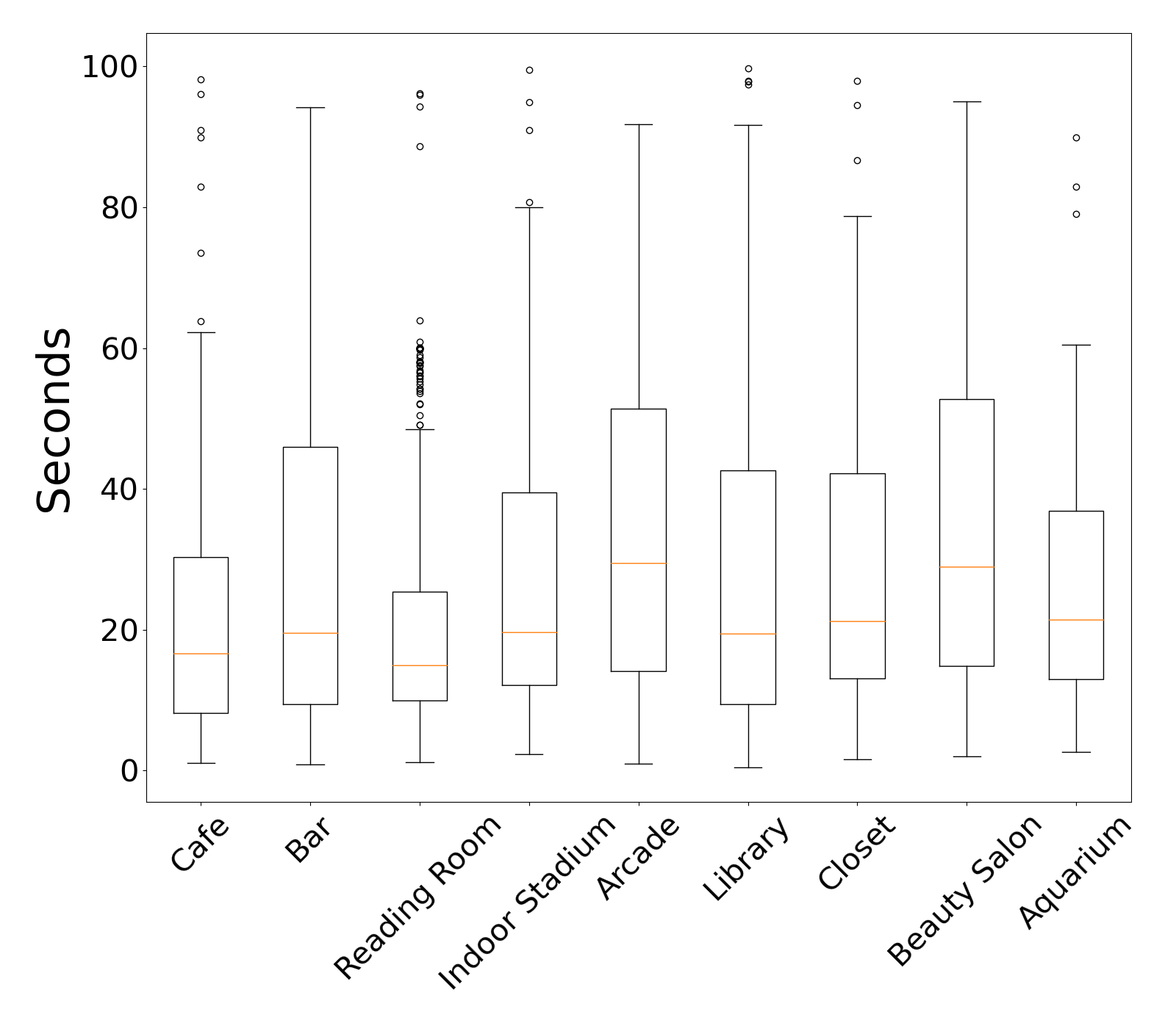}
   \caption{}
   \label{fig:insta_avg} 
\end{subfigure}
\hfill 
\begin{subfigure}[b]{0.47\textwidth}
   \includegraphics[width=1\linewidth]{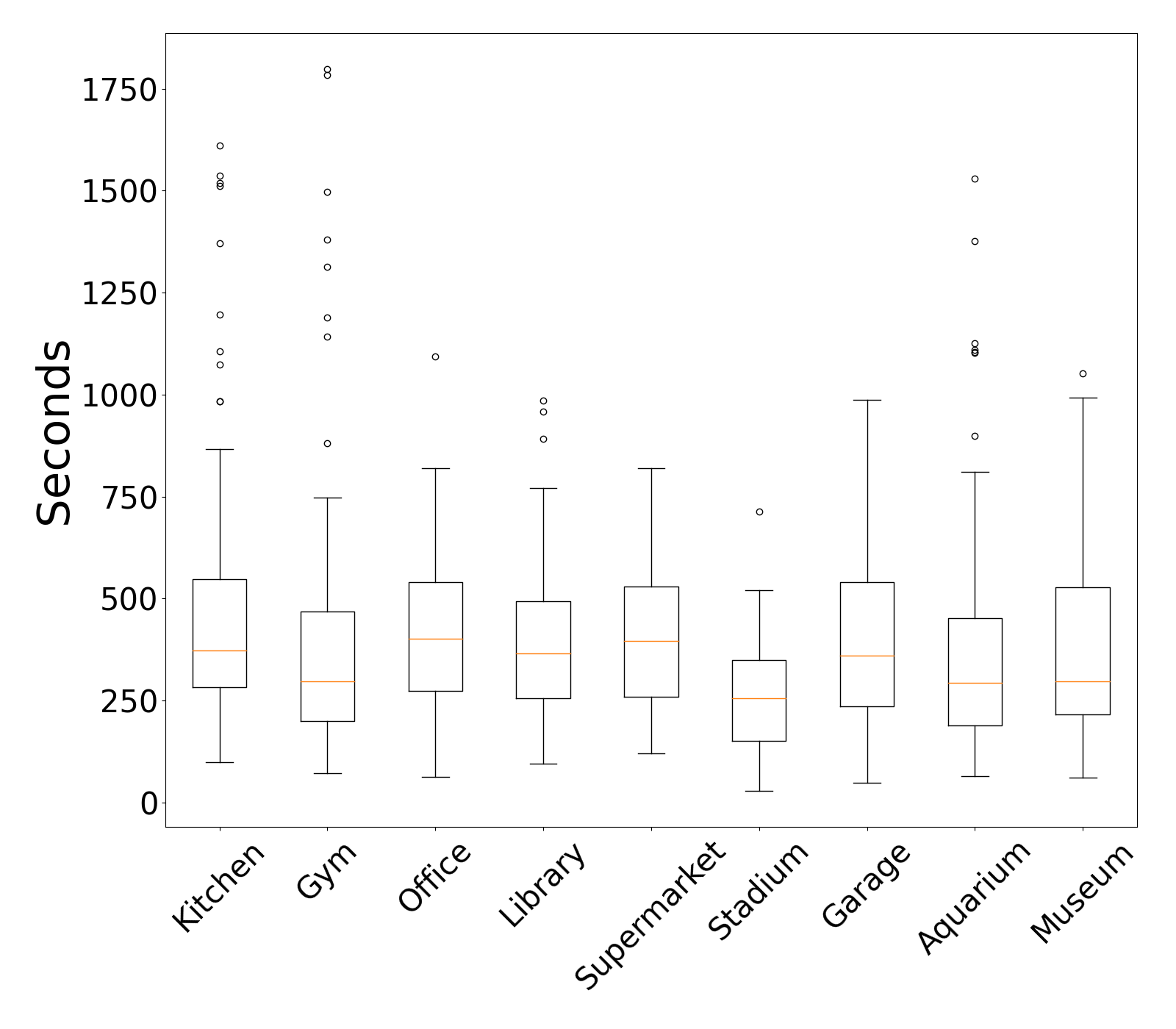}
   \caption{}
   \label{fig:yt_avg}
\end{subfigure}
\caption[Average Video Length]{Average length in seconds per class. (a) InstaIndoor dataset. (b) YouTubeIndoor.}
\label{fig:avg_len}
\end{figure}

Given the number of categories of this dataset and their variety, we select a subset consisting of relevant classes with respect to the tackled indoor scenes. The subset is selected such that it encompasses common indoor daily locations, comparable with the ones in the InstaIndoor dataset. It is worth mentioning that it was not possible to obtain videos corresponding to all classes present in the InstaIndoor dataset due to the fact that certain classes were not available in Youtube-8M. This subset is referred to as YouTubeIndoor throughout the rest of this paper, and consists of 9 classes, specifically: `Kitchen', `Gym', `Office', `Library', `Supermarket', `Stadium', `Garage', `Aquarium', and `Museum'. Each class contains 100 videos, containing English speech and images of high relevance to the respective scenes. Although these videos are longer, as seen in Figure \ref{fig:yt_avg}, and usually from a different angle or perspective, they cover similar scenes as the ones in our proposed InstaIndoor dataset.

This dataset is split in the same manner as the InstaIndoor dataset in terms of train:val:test, in order to ensure consistency. Both of the datasets proposed in this work, InstaIndoor and YouTubeIndoor, as well as the corresponding ground truth labels are publicly available for further research at \footnote{ Both datasets and corresponding pipeline code are available at \url{https://github.com/andreea-glavan/multimodal-audiovisual-scene-recognition}}.

\begin{figure}[h!]
\centering
\includegraphics[width=0.8\columnwidth]{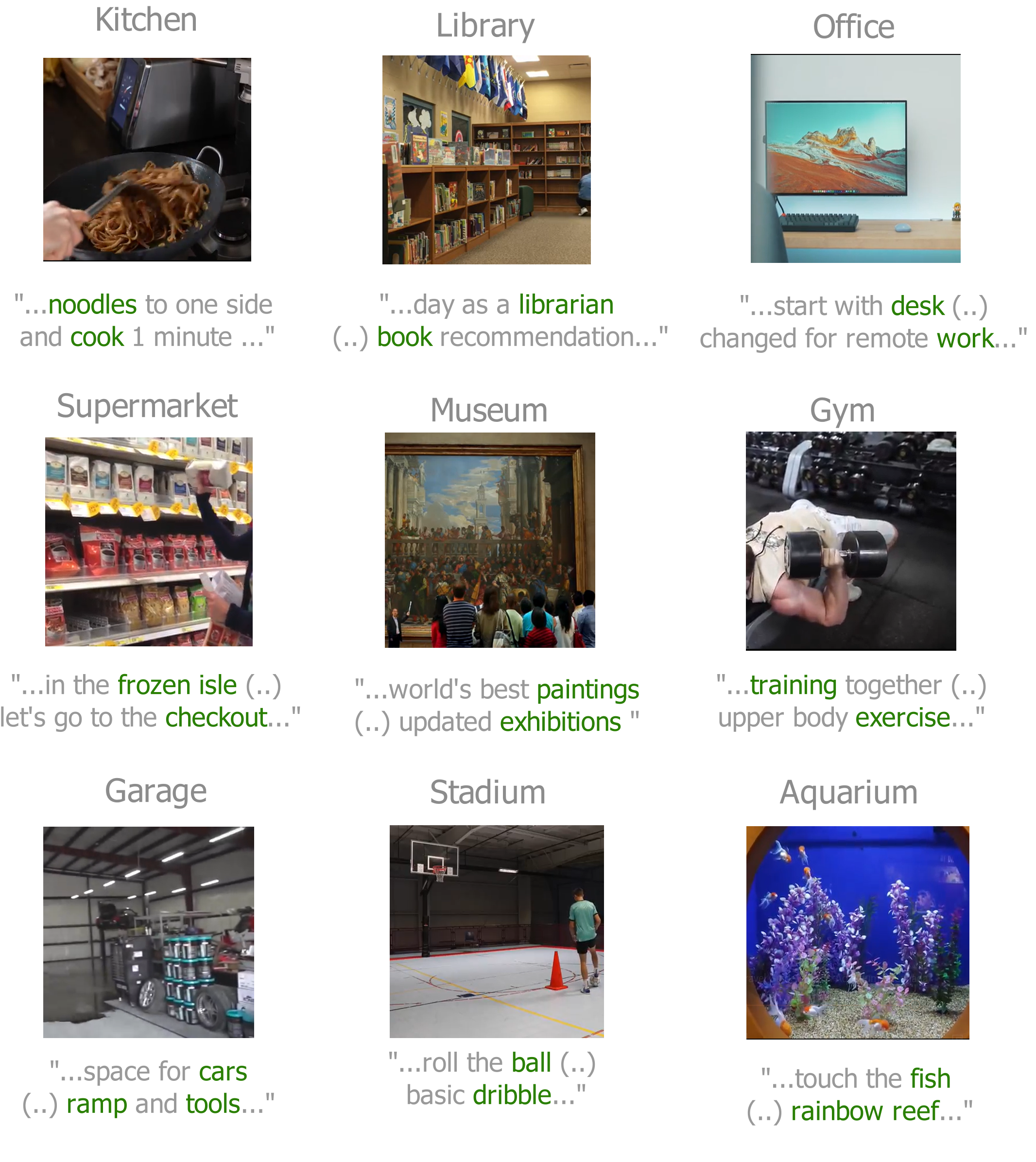}
\caption{Example images and transcribed text per class from YouTubeIndoor. Frame sampled randomly from the video; transcribed text selected to highlight key words.}
\label{fig:yt_img}
\end{figure}

\subsection{Experimental setup}

Next, we described our performed ablation study on different video data representation and the implementation details.

\subsubsection{Data representation}

In this work, we explore different ways of representing the visual and audio information. 

$\bullet$ \textbf{Visual}. Given an input video, we construct a sequence of frames by subsampling the video at the set rate of 10 frames per video with OpenCV \cite{opencv_library}. If an insufficient number of frames is reached and the video ends, the sequence is padded with blank images. Each frame in the sequence is resized to 64 x 64 and is ensured to be in RGB three channel format. 

We experiment with both frames as well as CNN features, extracted according to the process explained in Section \ref{sec:methodology}. We use VGG16 \cite{simonyan2014very} networks pre-trained on the ImageNet dataset and on the Places365 dataset to extract different global CNN features. These two specific networks were selected due to the semantic power of the pre-trained features they extract. The extracted descriptors are capable of capturing the scene or environment at both a local (object) and global (background) level.
    
\begin{itemize}
    \item Sequences of extracted frames of length 10 (referred to as Frames).
    \item Element-wise summed ImageNet feature vector values for each frame part of the sequence representing the video (referred to as ImgNFeat).
    \item Element-wise summed Places365 feature vector values for each frame part of the sequence representing the video (referred to as PlcFeat).
\end{itemize}
    
$\bullet$ \textbf{Audio, expressed as transcribed text.} Given an input text document generated according to the method presented in Section \ref{sec:methodology}, text embeddings can provide a representation in vector space. We experiment with a variety of text vectorization methods, ranging from classic approaches to state-of-the-art text embeddings.

\begin{itemize}
    \item Word2Vec \cite{mikolov2013efficient}, the most popular text embedding tool, is a recurrent neural network based model that operates at word level. As we deal with documents containing more than one word, using word embeddings requires further processing, in the form of sentence representations. We experiment with:
    \begin{itemize}
    \item Word2Vec representations per word as a list of embeddings, padded to length 100, a 100 dimension vector where each element is a one dimensional vector of length 100 corresponding to the respective word embedding values (referred to as W2V Pad).
    \item Element-wise summed value of Word2Vec embedding per word, as a one dimensional vector of length 100 corresponding to the embedding length (referred to as W2V Sum).
    \end{itemize}
    \item Count Vectorizer, a traditional Bag of Words representation, as used in \cite{zhang2010understanding}, transforms an input text body into a sparse matrix of token counts. We select the first 100 tokens from each text document, as after stopword removal they tend not to be over 100 words and we consider a vocabulary size of 20000 words (referred to as CountVect).

    \item SentenceBERT \cite{reimers2019sentence}, a transformer based model which operates at sentence level, is experimented with due to its capability of creating semantically meaningful embeddings which can be used to contextually compare larger text bodies. SentenceBERT was built on top of the state-of-the-art language models BERT \cite{devlin2018bert} and RoBERTa \cite{liu2019roberta}, and is capable of providing high accuracy with a reduced computation time. This tool represents a document as a one dimensional vector of length 768 (referred to as SentBERT).
\end{itemize}

\subsubsection{Single Modality Baseline Comparison}

As an additional experiment, we evaluate single modality approaches for the recognition of indoor scenes in InstaIndoor. To this end, we assess each of the features presented above with a corresponding model architecture.

For the visual features, we evaluate:

\begin{itemize}
    \item Frames: using a model based on a ConvLSTM with 512 units followed by a softmax activation layer.
    \item ImageNet features: using a dense, fully connected model. 
    \item Places features: using a dense, fully connected model. 
\end{itemize}

For the text features, we evaluate:

\begin{itemize}
    \item Count Vectorizer: using a model based on a LSTM with 512 units followed by a softmax activation layer. 
    \item Word2Vec Pad: using a model based on a LSTM with 512 units followed by a softmax activation layer.
    \item Word2Vec Sum: using a dense, fully connected model. 
    \item SentenceBERT: using a dense, fully connected model.
\end{itemize}


This experiment serves as a baseline for comparison with the multi-modal approach by utilising the same feature processing techniques in a singular feature context model.

\subsubsection{Implementation details}

Here, we detail the implementation and respective parameter values.

All models are trained for 20 epochs in batches of size 16, using the Adam optimizer algorithm \cite{kingma2014adam} with a learning rate $\alpha = 0.001 $, $\beta_1 = 0.9$, $\beta_2 = 0.999$, and $\epsilon = 1e-7$. The LSTM and ConvLSTM layers each contain 512 units, with the ConvLSTM using a kernel size of 3 $x$ 3. The fully connected dense layers vary from 128 to 512 units, with the layers prior to fusion containing 512 units, and the layers post fusion consisting of a series of 2 layers, with 256 and 128 units, respectively. The softmax layers contain 9 units, outputting feature vectors corresponding to the number of classes. The units per layer remain the same as mentioned above regardless of the multi-modal or single modality approach. An Early Stopping criterion \cite{caruana2001overfitting} is applied to the validation accuracy per epoch, ensuring the stopping of training after a lack of performance improvement, with a patience value of 10 epochs. Furthermore, during the end-to-end model training, we compute the categorical cross entropy loss. This measure can highlight how distinguishable two given probability distributions are from one another, defined as follows:
\begin{equation}
    cross\_entropy\_loss = \sum_{i=1}^{N} t_i log(s_i),
\end{equation}

where $N$ represents the number of classes, $t_i$ is the ground truth label, and $s_i$ is the model prediction.

All models and experiments were implemented using Python 3.7 \cite{10.5555/1593511} together with Keras \cite{chollet2015keras}, which provides an interface to TensorFlow \cite{tensorflow2015-whitepaper}.

\subsection{Evaluation} \label{subsection:eval}
The proposed methods are evaluated both quantitatively as well as qualitatively. The models are validated quantitatively by computing the Accuracy, Precision, Recall, and F1-Score. For all of these metrics, we compute both the macro and the weighted variants. The macro variant is calculated by dividing the class results by the total number of classes, whereas the weighted variant is calculated as the weighted average, thus multiplying the per-class results by the number of samples present in the respective class.

\begin{equation}
    Precision = \frac{TP}{TP+FP}
\end{equation}

\begin{equation}
    Recall  = \frac{TP}{TP+FN}
\end{equation}

\begin{equation}
    Accuracy = \frac{TP+TN}{TP+TN+FP+FN}
\end{equation}

\begin{equation}
    F1-Score = 2*\frac{Precision*Recall}{Precision+Recall}
\end{equation}
where $FP$, $TP$, $FN$, and $TN$ stand for false positives, true positives, false negatives, and true negatives, respectively.

The performance is also evaluated from a qualitative standpoint, by considering the following aspects: top classification results for relevant entries per class and class relations derived from confusion matrices.

\section{Results and Discussion}\label{sec:results}

In this section, the obtained results are presented and discussed. We present the quantitative results in table format and the qualitative results in the form of figures. The figures illustrate the performance for the configurations proposed in our experimental setup. We compare our model variants for both of the proposed datasets, thus evaluating every fusion strategy and feature combination. The results are organized based on the dataset under analysis.

\subsection{InstaIndoor Results}

\begin{table*}[b]
\centering
\large
\resizebox{\columnwidth}{!}{
\begin{tabular}{c|c|cccccc}
Descriptors & Accuracy    & \multicolumn{2}{c}{Precision}                                       & \multicolumn{2}{c}{Recall}                                          & \multicolumn{2}{c}{F1}                         \\
             &             & \multicolumn{1}{c|}{M}           & \multicolumn{1}{c|}{W}           & \multicolumn{1}{c|}{M}           & \multicolumn{1}{c|}{W}           & \multicolumn{1}{c|}{M}           & W           \\ \hline \hline
Frames       & 0.43 $\pm$ 0.01 & \multicolumn{1}{c|}{0.45 $\pm$ 0.02} & \multicolumn{1}{c|}{0.48 $\pm$ 0.01} & \multicolumn{1}{c|}{0.43 $\pm$ 0.00} & \multicolumn{1}{c|}{0.43 $\pm$ 0.00} & \multicolumn{1}{c|}{0.41 $\pm$ 0.01} & 0.43 $\pm$ 0.01 \\
ImgNFeat     & \textbf{0.61 $\pm$ 0.02} & \multicolumn{1}{c|}{0.61 $\pm$ 0.01} & \multicolumn{1}{c|}{0.63 $\pm$ 0.01} & \multicolumn{1}{c|}{0.59 $\pm$ 0.02} & \multicolumn{1}{c|}{0.60 $\pm$ 0.01} & \multicolumn{1}{c|}{0.61 $\pm$ 0.01} & 0.62 $\pm$ 0.01 \\
PlcFeat      & 0.60 $\pm$ 0.01 & \multicolumn{1}{c|}{0.64 $\pm$ 0.00} & \multicolumn{1}{c|}{0.64 $\pm$ 0.00} & \multicolumn{1}{c|}{0.58 $\pm$ 0.01} & \multicolumn{1}{c|}{0.60 $\pm$ 0.01} & \multicolumn{1}{c|}{0.59 $\pm$ 0.01} & 0.60 $\pm$ 0.00 \\ \hline
CountVect    & 0.11 $\pm$ 0.02 & \multicolumn{1}{c|}{0.01 $\pm$ 0.01} & \multicolumn{1}{c|}{0.01 $\pm$ 0.01} & \multicolumn{1}{c|}{0.11 $\pm$ 0.01} & \multicolumn{1}{c|}{0.11 $\pm$ 0.00} & \multicolumn{1}{c|}{0.02 $\pm$ 0.01} & 0.02 $\pm$ 0.00 \\
W2V Pad      & 0.17 $\pm$ 0.01 & \multicolumn{1}{c|}{0.02 $\pm$ 0.01} & \multicolumn{1}{c|}{0.03 $\pm$ 0.00} & \multicolumn{1}{c|}{0.11 $\pm$ 0.02} & \multicolumn{1}{c|}{0.17 $\pm$ 0.01} & \multicolumn{1}{c|}{0.03 $\pm$ 0.01} & 0.05 $\pm$ 0.01 \\
W2V Sum      &     0.15   $\pm$ 0.00      & \multicolumn{1}{c|}{ 0.02 $\pm$ 0.00}            & \multicolumn{1}{c|}{0.02 $\pm$ 0.00}            & \multicolumn{1}{c|}{0.11 $\pm$ 0.00}            & \multicolumn{1}{c|}{0.15 $\pm$ 0.00}            & \multicolumn{1}{c|}{0.03 $\pm$ 0.00}            &          0.04 $\pm$ 0.00   \\
SentBERT     &  0.13 $\pm$ 0.01          & \multicolumn{1}{c|}{0.01 $\pm$ 0.01}            & \multicolumn{1}{c|}{0/02 $\pm$ 0.00}            & \multicolumn{1}{c|}{0.11 $\pm$ 0.01}            & \multicolumn{1}{c|}{0.13 $\pm$ 0.01} & \multicolumn{1}{c|}{0.03 $\pm$ 0.01}            &          0.03 $\pm$ 0.00  
\end{tabular}
}
\caption{Results of all single modality experiment configurations on the InstagramIndoor dataset. Classification performance averaged across three runs. Best results are marked in bold.}
\label{tab:single-modality-res}
\end{table*}

The obtained baseline single-modality results on the InstaIndoor dataset are showcased in Table \ref{tab:single-modality-res}. It is immediate that the models using visual features outperform those using text features. This could be due to an insufficient amount of information being provided by the text in many videos, e.g., similar words being used in multiple scenes or not enough explanations being provided.

The best performing single modality approach uses ImageNet features, achieving 61\% accuracy. It is likely due to the pre-trained aspect of the features that this model is capable of best determining scenes. The Places features also perform well, however, they are surpassed by ImageNet, which is object focused. This may be related to the videos being shot as close to objects as opposed to landscapes. In terms of text features, the best performance is achieved by Word2Vec Pad, with 17\% accuracy. However, all text models tend to perform just slightly above random guessing (11\%).

\begin{table*}[b]
\centering
\large
\resizebox{\columnwidth}{!}{\begin{tabular}{l|l|l|c|c|c|c|c|c|c}
Text Ft. & Visual Ft. & Fusion & Acc. & \multicolumn{2}{c}{Precision} & \multicolumn{2}{c}{Recall} & \multicolumn{2}{c}{F1} \\
         &            &    &  & M          & W         & M          & W         & M          & W         \\ \hline \hline
CountVect & ImgNFeat & \multirow{8}{*}{Early} & 0.14 $\pm$ 0.01 & 0.02 $\pm$ 0.00 & 0.02 $\pm$ 0.01 & 0.11 $\pm$ 0.00 & 0.14 $\pm$ 0.01 & 0.02 $\pm$ 0.00 & 0.03 $\pm$ 0.01\\
CountVect & PlcFeat &  & 0.17 $\pm$ 0.00 & 0.02 $\pm$ 0.00 & 0.03 $\pm$ 0.00 & 0.11 $\pm$ 0.00 & 0.17 $\pm$ 0.00 & 0.03 $\pm$ 0.00 & 0.05 $\pm$ 0.00\\
W2V Pad     & ImgNFeat     &   &\textbf{0.70}	$\pm$ 0.02 &0.71 $\pm$ 0.02&	0.73$\pm$ 0.04&	0.68 $\pm$ 0.02&	0.70 $\pm$ 0.01&	0.69 $\pm$ 0.01 &	\textbf{0.70} $\pm$ 0.02                                  \\
W2V Pad     & PlcFeat       &   &   0.59 $\pm$ 0.02 & 0.62 $\pm$ 0.00 & 0.64 $\pm$ 0.01 & 0.57 $\pm$ 0.03 & 0.59 $\pm$ 0.02 & 0.58 $\pm$ 0.00 & 0.59 $\pm$ 0.01 \\
W2V Sum     & PlcFeat       &   & 0.55 $\pm$ 0.00                    & 0.54  $\pm$ 0.01                                & 0.55 $\pm$ 0.00                                     & 0.51     $\pm$ 0.01                          & 0.55 $\pm$ 0.00                                 & 0.51 $\pm$ 0.02                               & 0.53 $\pm$ 0.00                                  \\
W2V Sum     & ImgNFeat  &   & \textbf{0.69} $\pm$ 0.01 &	0.70 $\pm$ 0.02 &	0.71 $\pm$ 0.01 &	0.68 $\pm$ 0.01 &	0.69 $\pm$ 0.01&	0.68 $\pm$ 0.02 &	0.69 $\pm$ 0.01                                  \\
SentBERT     & ImgNFeat      &   & 0.66 $\pm$ 0.01&	0.68 $\pm$ 0.02&	0.70 $\pm$ 0.01 &	0.64 $\pm$ 0.02 &	0.66 $\pm$ 0.02 &	0.63 $\pm$ 0.02 &	0.66 $\pm$ 0.01                                  \\
SentBERT     & PlcFeat        &   & 0.55  $\pm$ 0.02                        & 0.50     $\pm$ 0.00                              & 0.53                           $\pm$ 0.01          & 0.50 $\pm$ 0.00                                & 0.53         $\pm$ 0.01                         & 0.49 $\pm$ 0.01                              & 0.51 $\pm$ 0.01                                  \\
\hline
CountVect & Frames            & \multirow{12}{*}{Joint}  & 0.46         $\pm$   0.05             & 0.45  $\pm$   0.02                                  & 0.47  $\pm$    0.01                                    & 0.45 $\pm$   0.01                               & 0.46  $\pm$   0.01                                 & 0.44   $\pm$   0.02                             & 0.46     $\pm$   0.01                              \\

CountVect & ImgNFeat      &   & \textbf{0.68}  $\pm$   0.02                         & 0.67           $\pm$   0.01                        & 0.68  $\pm$   0.00                                    & 0.66  $\pm$   0.02                              & 0.68  $\pm$   0.01                                 & 0.66  $\pm$   0.01                              & \textbf{0.68}  $\pm$   0.01                                 \\
CountVect & PlcFeat        &   & 0.61  $\pm$   0.01                         & 0.61                $\pm$        0.01              & 0.62              $\pm$   0.02                        & 0.59            $\pm$   0.02                    & 0.61  $\pm$   0.01                                  & 0.59   $\pm$   0.01                             & 0.60     $\pm$   0.01                               \\
W2V Pad     & Frames            &   & 0.36         $\pm$   0.01                 & 0.42     $\pm$   0.00                              & 0.43  $\pm$   0.00                                    & 0.34   $\pm$   0.01                             & 0.36      $\pm$   0.00                             & 0.33     $\pm$   0.01                           & 0.35       $\pm$   0.00                            \\
W2V Pad     & ImgNFeat      &   & 0.67     $\pm$   0.02                     & 0.66         $\pm$   0.01                          & 0.68  $\pm$   0.01                                    & 0.66   $\pm$   0.01                             & 0.67  $\pm$   0.00                                 & 0.66     $\pm$   0.01                           & 0.67               $\pm$   0.01                    \\
W2V Pad     & PlcFeat        &   & 0.57        $\pm$   0.01                  & 0.57     $\pm$   0.00                              & 0.59   $\pm$   0.00                                   & 0.56  $\pm$   0.01                              & 0.57  $\pm$   0.01                                 & 0.56     $\pm$   0.01                           & 0.58       $\pm$   0.00                            \\
W2V Sum     & Frames            &   & 0.47     $\pm$   0.00                     & 0.47         $\pm$   0.00                          & 0.50  $\pm$   0.00                                     & 0.48           $\pm$   0.01                     & 0.47     $\pm$   0.00                              & 0.46  $\pm$   0.00                              & 0.47  $\pm$   0.00                                 \\
W2V Sum     & ImgNFeat      &   & \textbf{0.68}  $\pm$   0.01                         & 0.68   $\pm$   0.01                                & 0.69                      $\pm$   0.01                & 0.66  $\pm$   0.00                              & 0.68            $\pm$   0.00                       & 0.66  $\pm$   0.01                              & 0.68     $\pm$   0.00                              \\
W2V Sum     & PlcFeat        &   & 0.61     $\pm$   0.01                     & 0.61         $\pm$   0.00                          & 0.62  $\pm$   0.00                                    & 0.59          $\pm$   0.01                      & 0.61          $\pm$   0.00                         & 0.60   $\pm$   0.01                              & 0.61                $\pm$   0.00                   \\
SentBERT     & Frames            &   & 0.46      $\pm$   0.03                    & 0.46     $\pm$   0.02                              & 0.48  $\pm$   0.01                                    & 0.45  $\pm$   0.01                              & 0.46  $\pm$   0.00                                 & 0.44     $\pm$   0.02                           & 0.46   $\pm$   0.01                                \\
SentBERT     & ImgNFeat      &   & 0.64   $\pm$   0.01                       & 0.66    $\pm$   0.00                               & 0.67    $\pm$   0.00                                  & 0.62              $\pm$   0.01                  & 0.64       $\pm$   0.01                            & 0.63      $\pm$   0.01                          & 0.65                       $\pm$   0.00            \\
SentBERT     & PlcFeat        &   & 0.57  $\pm$   0.00                        & 0.59       $\pm$   0.00                            & 0.61   $\pm$   0.00                                   & 0.57            $\pm$   0.00                    & 0.57        $\pm$   0.00                           & 0.56      $\pm$   0.00                          & 0.58        $\pm$   0.00                           \\
\hline
CountVect & Frames            & \multirow{12}{*}{Late}   & 0.13 $\pm$ 0.00&	0.01 $\pm$ 0.00&	0.02 $\pm$ 0.00&	0.11 $\pm$ 0.00&	0.13 $\pm$ 0.00&	0.02 $\pm$ 0.00&	0.03 $\pm$ 0.00 \\
CountVect & ImgNFeat      &    & 0.40 $\pm$ 0.01&	0.28 $\pm$ 0.01&	0.30 $\pm$ 0.00&	0.34 $\pm$ 0.02&	0.40 $\pm$ 0.00&	0.27 $\pm$ 0.01&	0.32  $\pm$ 0.01\\
CountVect & PlcFeat        &    & 0.40 $\pm$  0.02&	0.43 $\pm$  0.00&	0.46 $\pm$  0.00&	0.41 $\pm$  0.01&	0.40 $\pm$  0.01&	0.35 $\pm$  0.01&	0.34 $\pm$  0.00\\
W2V Pad     & Frames            &    & 0.13 $\pm$ 0.01&	0.03 $\pm$ 0.01&	0.04 $\pm$ 0.01&	0.11 $\pm$ 0.01&	0.13 $\pm$ 0.00&	0.03 $\pm$ 0.01&	0.03 $\pm$ 0.00\\
W2V Pad     & ImgNFeat      &    & 0.54 $\pm$ 0.01&	0.51 $\pm$ 0.01&	0.54 $\pm$ 0.00&	0.50 $\pm$ 0.01&	0.54 $\pm$ 0.00&	0.43 $\pm$ 0.01&	0.47 $\pm$ 0.00\\
W2V Pad     & PlcFeat        &    & 0.23 $\pm$  0.00&	0.23 $\pm$  0.00&	0.32 $\pm$  0.01&	0.33 $\pm$  0.00&	0.23 $\pm$  0.01&	0.23 $\pm$  0.00 & 0.21 $\pm$ 0.01 \\
W2V Sum     & Frames            &    & 0.21 $\pm$ 0.00&	0.09 $\pm$ 0.01&	0.12 $\pm$ 0.00&	0.16 $\pm$ 0.00&	0.21 $\pm$ 0.00&	0.12 $\pm$ 0.00&	0.15 $\pm$ 0.00\\
W2V Sum     & ImgNFeat      &    &0.18  $\pm$ 0.00&	0.23  $\pm$ 0.01&	0.28  $\pm$ 0.00&	0.17  $\pm$ 0.00&	0.18  $\pm$ 0.00&	0.10  $\pm$ 0.00&	0.12  $\pm$ 0.00\\
W2V Sum     & PlcFeat        &    & 0.34  $\pm$  0.01&	0.23  $\pm$  0.01&	0.27  $\pm$  0.00&	0.30  $\pm$  0.01&	0.34  $\pm$  0.00&	0.24  $\pm$  0.01&	0.27  $\pm$  0.00\\
SentBERT     & Frames            &    & 0.18 $\pm$ 0.01&	0.16 $\pm$ 0.02&	0.19 $\pm$ 0.01&	0.17 $\pm$ 0.00&	0.18 $\pm$ 0.00&	0.12 $\pm$ 0.01&	0.13 $\pm$ 0.01\\
SentBERT     & ImgNFeat      &    & 0.19 $\pm$ 0.01&	0.24 $\pm$ 0.01&	0.27 $\pm$ 0.00&	0.14 $\pm$ 0.02&	0.19 $\pm$ 0.00&	0.09 $\pm$ 0.01&	0.11 $\pm$ 0.00\\
SentBERT     & PlcFeat        &    & 0.27 $\pm$ 0.02&	0.18 $\pm$ 0.01&	0.20 $\pm$ 0.01&	0.21 $\pm$ 0.02&	0.27 $\pm$ 0.00&	0.16 $\pm$ 0.02&	0.19 $\pm$ 0.01\\

\end{tabular}}
\caption{Results of all experiment configurations on the InstagramIndoor dataset. Classification performance, averaged across three runs. Best results are marked in bold.}
\label{tab:insta-res-late}
\end{table*}

We present the obtained results of our ablation study over our newly introduced InstaIndoor dataset in Table \ref{tab:insta-res-late}. As we can observe, the best performing model is based on Early fusion and uses ImageNet Sum as visual features and Word2Vec Padded as text features, achieving 70\% accuracy. The second best model shares similar features, with the only exception being the text feature as Word2Vec Sum, which achieves 69\% accuracy. Overall, all top 5 performing models, marked in bold in Table \ref{tab:insta-res-late}, use ImageNet Sum features but vary in terms of text processing and/or fusion strategy. 

We observe that overall early and joint fusion slightly out-perform late fusion in terms of quantitative metrics, with early fusion having overall the best results regardless of the combination of features used. The joint fusion results are similar to those of early fusion. However, these results tend to be slightly lower, at most 10\% in terms of average accuracy. The majority of late fusion results are comparable with the other two architectures, however, certain features perform poorer in this case, such as Frames + Count Vectorizer with 13\% average accuracy. However, these results are still better than random guessing: $\frac{1}{9 \: classes} = 0.11$ accuracy. The CNN features obtain some of the best results in this case as well, with 54\% average accuracy alongside Word2Vec Pad text feature.

With respect to text features, we notice that the Count Vectorizer performs poorly, in the worst case reaching 13\% accuracy (Late fusion with Frames as visual features). The best performing text features tend to be the Word2Vec Pad and Word2Vec Sum. It may be the case that SentenceBERT features perform slightly poorer due to minor differences in speech and lack of key words spoken. Furthermore, in terms of visual features, the best performing models use CNN features as opposed to frames. Given the fact that CNNs are pre-trained on large datasets depicting a wide range of scenarios, it is likely that these models capture more information than the ConvLSTM can in the limited training of 20 epochs. The ImageNet features appear to lead to better results than the Places365 features in most cases, which may be related to the way in which videos are filmed on mobile devices, i.e., close-up shots, object focused amateur videos, etc.

Compared to the single modality baseline results, we notice that the results indicate a minimum of a 10\% increase. By connecting visual and text features, we obtain better performance than either descriptors can on their own. Furthermore, it is notable that the best performing features tend to be the same regardless of whether the system is single or multi-modal, i.e., ImageNet visual features tend to lead to the best performance. The text features appear to perform more similarly to one another, however, the Word2Vec features appear to retain their slight edge over the others, as they do in the single modality approach as well.

\begin{figure}[h!]
\centering
\includegraphics[width=0.6\columnwidth]{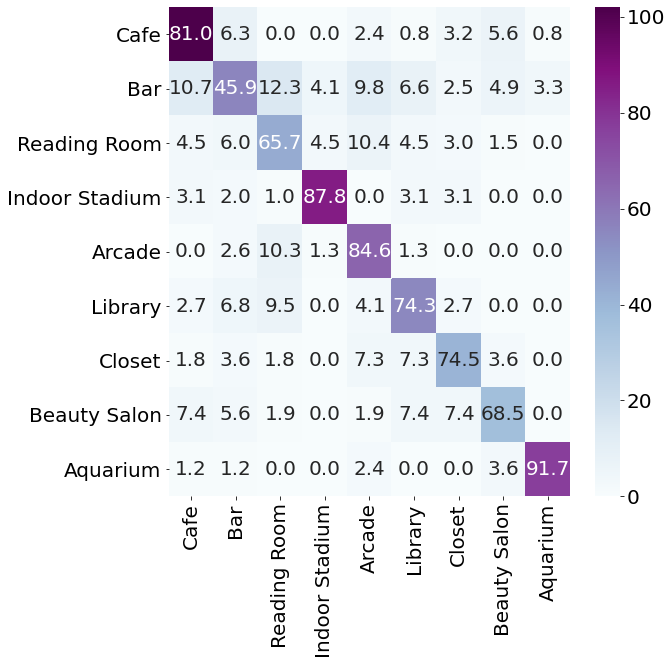}
\caption{Confusion Matrix for the best performing model on the InstaIndoor dataset: Early fusion of ImageNet Sum visual feature and Word2Vec Pad text features.}
\label{fig:insta_cm}
\end{figure}

From a qualitative perspective, we notice from the confusion matrix in Figure \ref{fig:insta_cm} that there is a high level of confusion for classes such as `Closet' or `Beauty Salon', which may be attributed to the high intra-class variance and the low number of specialized terms spoken in such videos. Furthermore, the `Reading Room' class has the lowest percentage of true positive predictions, specifically 65.7. This class especially suffers from high intra-class variance due to factors such as decor preferred by individuals. It also faces problems with respect to low inter-class variance with classes such as `Arcade' or `Library' due to either lighting or objects present.

\begin{figure}[h!]
\centering
\includegraphics[width=0.9\columnwidth]{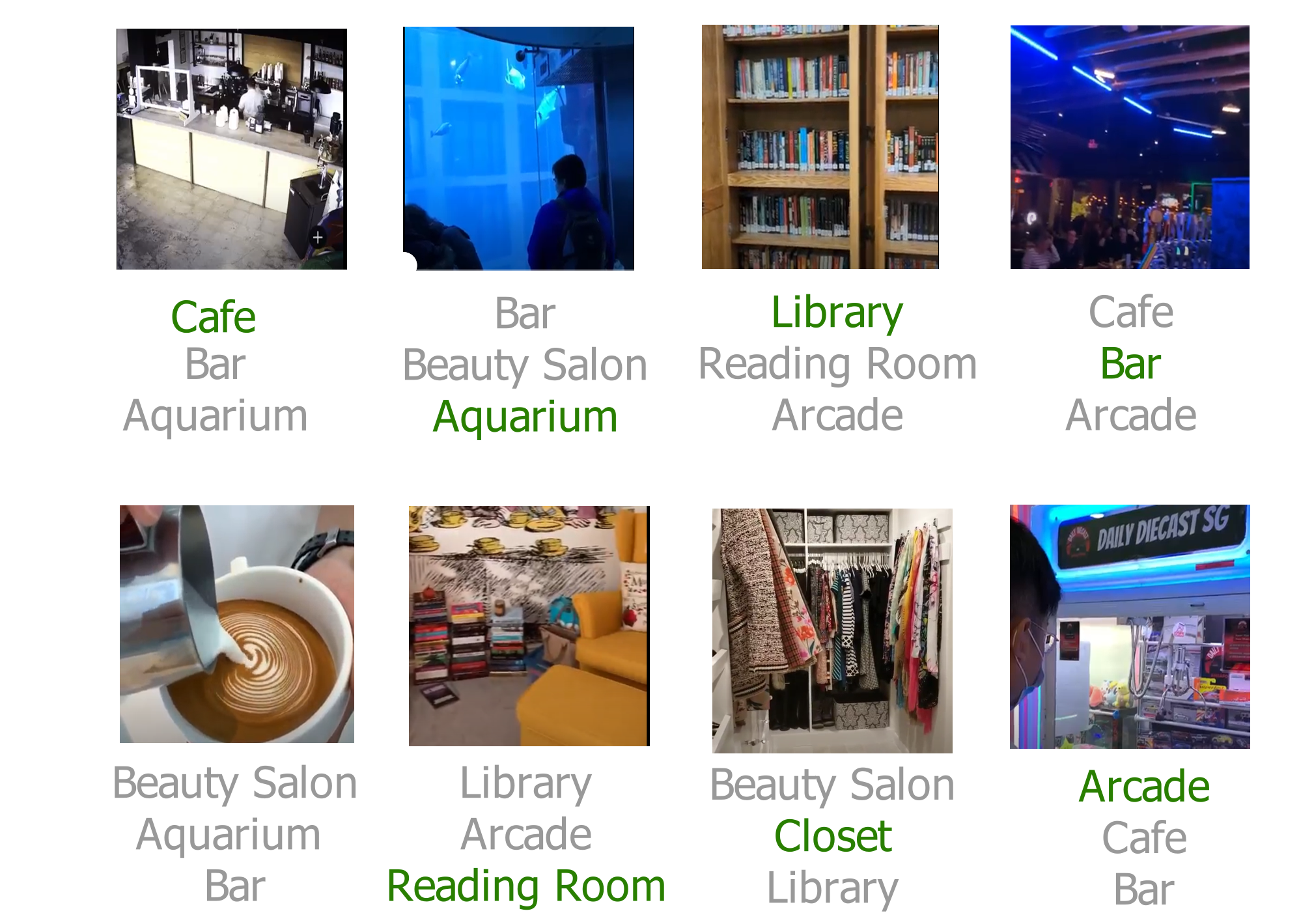}
\caption{Example top predictions of the best performing model on the InstaIndoor dataset. Frames sampled such that they are representative to the corresponding scene.}
\label{fig:insta_pred}
\end{figure}

When looking at results for sample videos in Figure \ref{fig:insta_pred}, we notice that the `Bar' class seems to be confused for `Cafe' and other classes, such as `Reading Room', are confused for `Library'. A possible reason for this could be the shared objects between certain classes. For instance, glasses are present in both `Cafe' and `Bar'; books are present in both `Reading Room' and `Library'. Moreover, we can observe an example of confusion due to lighting colours in Pictures 2 and 4 from Figure \ref{fig:insta_pred}, where both share a blue light despite belonging to different classes. 

Furthermore, videos focused on only one object appear to not be classified correctly. We encounter this phenomenon multiple times during our qualitative evaluation, often with coffee cups as illustrated in the first image of the second row in Figure \ref{fig:insta_pred}. A possible explanation for this happening could be a relatively low number of such close up videos being present in the training set, with the focus falling on the background or other objects in the respective scene. Another possibility is the overlapping of such objects between different classes. For instance, a coffee cup could be present in both a `Cafe' or a `Bar', perhaps a `Library'. Therefore, the low inter-class variance would lead to confusion when the focus falls on mainly one object.

\subsection{YouTubeIndoor Results}

The quantitative results of our study on the YouTubeIndoor subset are illustrated in Table \ref{tab:yt-res-late}. In this case, the best performing model uses a Joint fusion of ImageNet Sum visual features and Count Vectorizer text features, with 74\% accuracy. It is notable that, despite the better performance, fewer model configurations achieve such results on this dataset. While on InstaIndoor the performances of the top 5 configurations were comparable, in this case only the top 2 reach above 70\% accuracy. This may be due to the greater length and amount of editing present in YouTube videos. Furthermore, editing cuts from one scene to another can cause confusion.

Certain performance elements remain the same as in the case of the InstaIndoor dataset, such as the better performance of Early and Joint fusion. In this case, the Late fusion performs more poorly, with the best late fusion result being 36\% average accuracy, for Count Vectorizer and Places Sum. Similarly to the InstaIndoor results, we notice the better performance of pre-trained CNN features, with both ImageNet and Places features. It is likely that the Places features perform better on this dataset than on InstaIndoor due to the different angles of filming and increased camera distance. As was the case with InstaIndoor, the pre-trained weights play a significant role in scene recognition, especially when considering social media recording features.

\begin{figure}[h!]
\centering
\includegraphics[width=0.6\columnwidth]{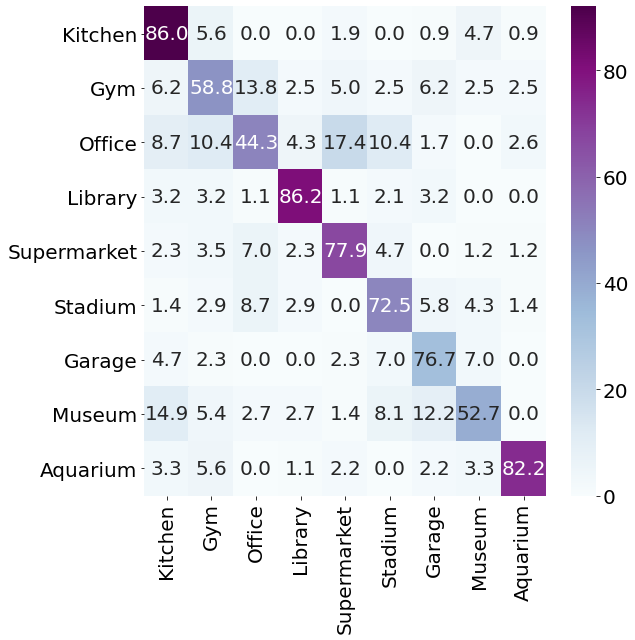}
\caption{Confusion Matrix for the best performing model on the YouTubeIndoor subset: Joint fusion of ImageNet Sum visual feature and Count Vectorizer text features.}
\label{fig:yt_cm}
\end{figure}

\begin{table*}
\centering
\large
\resizebox{\columnwidth}{!}{\begin{tabular}{l|l|l|c|c|c|c|c|c|c}
Text Ft. & Visual Ft. & Fusion & Acc. & \multicolumn{2}{c}{Precision} & \multicolumn{2}{c}{Recall} & \multicolumn{2}{c}{F1} \\
         &            &    &  & M          & W         & M          & W         & M          & W         \\ \hline \hline
CountVect & ImgNFeat  &\multirow{8}{*}{Early}   & 0.21 $\pm$ 0.00 & 0.03 $\pm$ 0.00 & 0.04 $\pm$ 0.00 & 0.14 $\pm$ 0.00 & 0.21 $\pm$ 0.01 & 0.05 $\pm$ 0.01& 0.07 $\pm$ 0.00 \\
CountVect & PlcFeat  &   & 0.21 $\pm$ 0.00 & 0.03 $\pm$ 0.00 & 0.04 $\pm$ 0.01 & 0.14 $\pm$ 0.01 & 0.21 $\pm$ 0.01 & 0.05 $\pm$ 0.00 & 0.07 $\pm$ 0.00 \\
W2V Pad     & ImgNFeat      &   & 0.20 $\pm$ 0.02 &	0.10 $\pm$ 0.00 &	0.10 $\pm$ 0.01 &	0.18 $\pm$ 0.02 &	0.19 $\pm$ 0.01 &	0.11 $\pm$ 0.01 &	0.12    $\pm$ 0.01                              \\
W2V Pad     & PlcFeat        &   & 0.61 $\pm$ 0.02 &	0.60 $\pm$ 0.00 &	0.63 $\pm$ 0.01 &	0.60 $\pm$ 0.02 &	0.60 $\pm$ 0.02 &	0.59 $\pm$ 0.01 &	0.59 $\pm$ 0.01                                  \\
W2V Sum     & ImgNFeat  &   & 0.40 $\pm$ 0.01 &	0.42 $\pm$ 0.02 &	0.46 $\pm$ 0.01 &	0.40 $\pm$ 0.00 &	0.40 $\pm$ 0.00 &	0.38 $\pm$ 0.01 &	0.40 $\pm$ 0.01                                  \\
W2V Sum     & PlcFeat        &   & 0.60 $\pm$ 0.03	&0.60 $\pm$ 0.00 &	0.60 $\pm$ 0.01&	0.62 $\pm$ 0.01 &	0.65 $\pm$ 0.02 &	0.59 $\pm$ 0.01 &	0.59 $\pm$ 0.01    \\
SentBERT     & ImgNFeat      &   & 0.60 $\pm$ 0.04	&0.64 $\pm$ 0.02 &	0.65 $\pm$ 0.00&	0.60 $\pm$ 0.03&	0.60 $\pm$ 0.02 &	0.59 $\pm$ 0.02 &	0.59 $\pm$ 0.02                                \\
SentBERT     & PlcFeat        &   & 0.65 $\pm$ 0.01 &	0.63 $\pm$ 0.00 &	0.66 $\pm$ 0.02 &	0.63 $\pm$ 0.01 &	0.65 $\pm$ 0.01 &	0.62 $\pm$ 0.01 &	0.65 $\pm$ 0.01                                  \\

\hline
CountVect & Frames            & \multirow{12}{*}{Joint}  & 0.16 $\pm$ 0.01 &	0.11$\pm$ 0.01&	0.10 $\pm$ 0.01&	0.16$\pm$ 0.00&	0.16 $\pm$ 0.00&	0.08 $\pm$ 0.01&	0.07 $\pm$ 0.01                                \\
CountVect & ImgNFeat      &   & \textbf{0.74} $\pm$ 0.01 &	0.73 $\pm$ 0.00&	0.76$\pm$ 0.00&	0.74$\pm$ 0.01&	0.74$\pm$ 0.00&	0.73 $\pm$ 0.01&	\textbf{0.74} $\pm$ 0.00                       \\
CountVect & PlcFeat        &   & 0.64 $\pm$ 0.02 &	0.63 $\pm$ 0.01&	0.64 $\pm$ 0.01&	0.63 $\pm$ 0.01&	0.64 $\pm$ 0.00&	0.63 $\pm$ 0.01&	0.64 $\pm$ 0.00                          \\

W2V Pad     & Frames            &   & 0.21 $\pm$ 0.00&	0.03 $\pm$ 0.01&	0.04 $\pm$ 0.00&	0.14 $\pm$ 0.00&	0.21 $\pm$ 0.00&	0.05 $\pm$ 0.00&	0.07     $\pm$ 0.00                             \\
W2V Pad     & ImgNFeat      &   & 0.22 $\pm$ 0.01&	0.21 $\pm$ 0.00&	0.23 $\pm$ 0.00&	0.24 $\pm$ 0.01&	0.23 $\pm$ 0.00&	0.20 $\pm$ 0.01&	0.19   $\pm$ 0.00                             \\
W2V Pad     & PlcFeat        &   & 0.40 $\pm$ 0.01&	0.42 $\pm$ 0.01&	0.46 $\pm$ 0.00&	0.40 $\pm$ 0.01&	0.40 $\pm$ 0.00&	0.38 $\pm$ 0.01&	0.40  $\pm$ 0.00                                \\

W2V Sum     & Frames            &   & 0.12 $\pm$ 0.02 &	0.11 $\pm$ 0.01 &	0.12 $\pm$ 0.01&	0.06 $\pm$ 0.01&	0.03 $\pm$ 0.01&	0.04 $\pm$ 0.01&	0.05  $\pm$ 0.01                                \\
W2V Sum     & ImgNFeat      &   & 0.22 $\pm$ 0.01 &	0.24 $\pm$ 0.00 &	0.27 $\pm$ 0.00 &	0.23 $\pm$ 0.01&	0.22 $\pm$ 0.01&	0.21 $\pm$ 0.01&	0.22 $\pm$ 0.00                                 \\

W2V Sum     & PlcFeat        &   & 0.24 $\pm$ 0.01&	0.28 $\pm$ 0.00 &	0.24 $\pm$ 0.00&	0.20 $\pm$ 0.01&	0.21 $\pm$ 0.00&	0.21 $\pm$ 0.01&	0.22 $\pm$ 0.00                                 \\

SentBERT     & Frames            &   & 0.12 $\pm$ 0.00&	0.03 $\pm$ 0.00&	0.04 $\pm$ 0.00&	0.11 $\pm$ 0.00&	0.12 $\pm$ 0.00 &	0.04 $\pm$ 0.00&	0.05 $\pm$ 0.00                                  \\
SentBERT     & ImgNFeat      &   & \textbf{0.69} $\pm$ 0.01 &	0.68 $\pm$ 0.01 &	0.70 $\pm$ 0.01&	0.69 $\pm$ 0.00&	0.69 $\pm$ 0.00 &	0.68 $\pm$ 0.01 &	\textbf{0.69}    $\pm$ 0.01                              \\

SentBERT     & PlcFeat        &   & 0.61 $\pm$ 0.01&	0.60 $\pm$ 0.01 &	0.62 $\pm$ 0.00&	0.60 $\pm$ 0.01&	0.61 $\pm$ 0.00 &	0.60 $\pm$ 0.01 &	0.61 $\pm$ 0.01                                 \\

\hline
CountVect & Frames            & \multirow{12}{*}{Late}   & 0.21 $\pm$ 0.01&		0.19 $\pm$ 0.01&	0.17 $\pm$ 0.01&	0.21 $\pm$ 0.00&	0.23 $\pm$ 0.00&	0.12 $\pm$ 0.00&	0.13 $\pm$ 0.00 \\
CountVect & ImgNFeat      &    & 0.35 $\pm$ 0.01&	0.35 $\pm$ 0.00&	0.33 $\pm$ 0.01&	0.40 $\pm$ 0.01&	0.35 $\pm$ 0.01&	0.33 $\pm$ 0.01&	0.30 $\pm$ 0.01\\
CountVect & PlcFeat        &    & 0.36 $\pm$ 0.01	&	0.30 $\pm$ 0.01&	0.28 $\pm$ 0.01&	0.42 $\pm$ 0.00&	0.38 $\pm$ 0.00&	0.32 $\pm$ 0.01&	0.28 $\pm$ 0.00 \\
W2V Pad     & Frames            &    & 0.14 $\pm$ 0.01	&	0.02 $\pm$ 0.01	&0.03 $\pm$ 0.01&	0.15 $\pm$ 0.01&	0.14 $\pm$ 0.00&	0.04 $\pm$ 0.01&	0.05 $\pm$ 0.00\\
W2V Pad     & ImgNFeat      &    & 0.18 $\pm$  0.00&	0.21 $\pm$  0.00&	0.23 $\pm$  0.00&	0.17 $\pm$  0.00&	0.18 $\pm$  0.00&	0.09 $\pm$  0.00&	0.11 $\pm$  0.00\\
W2V Pad     & PlcFeat        &    &    0.15 $\pm$ 0.00&	0.02 $\pm$ 0.00&	0.03 $\pm$ 0.00&	0.11 $\pm$ 0.00&	0.15 $\pm$ 0.00&	0.03 $\pm$ 0.00&	0.04 $\pm$ 0.00\\
W2V Sum     & Frames            &    &  0.19 $\pm$  0.00	&	0.11 $\pm$ 0.01&	0.14 $\pm$  0.00&	0.19 $\pm$  0.00&	0.21 $\pm$  0.00&	0.17 $\pm$  0.00&	0.18 $\pm$  0.00\\
W2V Sum     & ImgNFeat      &    & 0.27 $\pm$ 0.01&	0.11 $\pm$ 0.01&	0.15 $\pm$ 0.00&	0.27 $\pm$ 0.01&	0.25 $\pm$ 0.01&	0.14 $\pm$ 0.01&	0.16 $\pm$ 0.01\\
W2V Sum     & PlcFeat        &    & 0.31 $\pm$  0.01&	0.28 $\pm$  0.01&	0.27 $\pm$  0.01&	0.32 $\pm$  0.01&	0.36 $\pm$  0.00&	0.29 $\pm$  0.01&	0.27 $\pm$  0.00\\
SentBERT     & Frames            &    & 0.24 $\pm$ 0.02&	0.09 $\pm$ 0.01&	0.12 $\pm$ 0.00&	0.16 $\pm$ 0.00&	0.21 $\pm$ 0.00&	0.12 $\pm$ 0.00&	0.15 $\pm$ 0.00\\
SentBERT     & ImgNFeat      &    & 0.15 $\pm$ 0.02&	0.11 $\pm$ 0.01&	0.10 $\pm$ 0.01&	0.18 $\pm$ 0.02&	0.15 $\pm$ 0.01&	0.09 $\pm$ 0.01&	0.08 $\pm$ 0.01\\
SentBERT     & PlcFeat        &    & 0.13 $\pm$  0.01&	0.05 $\pm$  0.01&	0.05 $\pm$  0.00&	0.15 $\pm$  0.01&	0.13 $\pm$  0.01&	0.05 $\pm$  0.01&	0.05 $\pm$  0.01
\end{tabular}}
\caption{Results of all experiment configurations on YouTubeIndoor. Classification performance, averaged across three runs. Best results are in bold.}
\label{tab:yt-res-late}
\end{table*}

Unlike with the InstaIndoor dataset, the best performing text feature, in this case, is the Count Vectorizer. A possible explanation for this is that due to the greater video length and the fact that videos are often topic or tutorial focused, specialized terms appear more often. 
Thus, the specialized term to class correlation can be better captured by the Count Vectorizer as it emphasizes terms rarely encountered outside certain classes.
Another notable text feature is the SentenceBERT embedding, which performs significantly better than the Word2Vec embeddings, likely due to the relevance of the words spoken in tutorials.

\begin{figure}[h!]
\centering
\includegraphics[width=0.8\columnwidth]{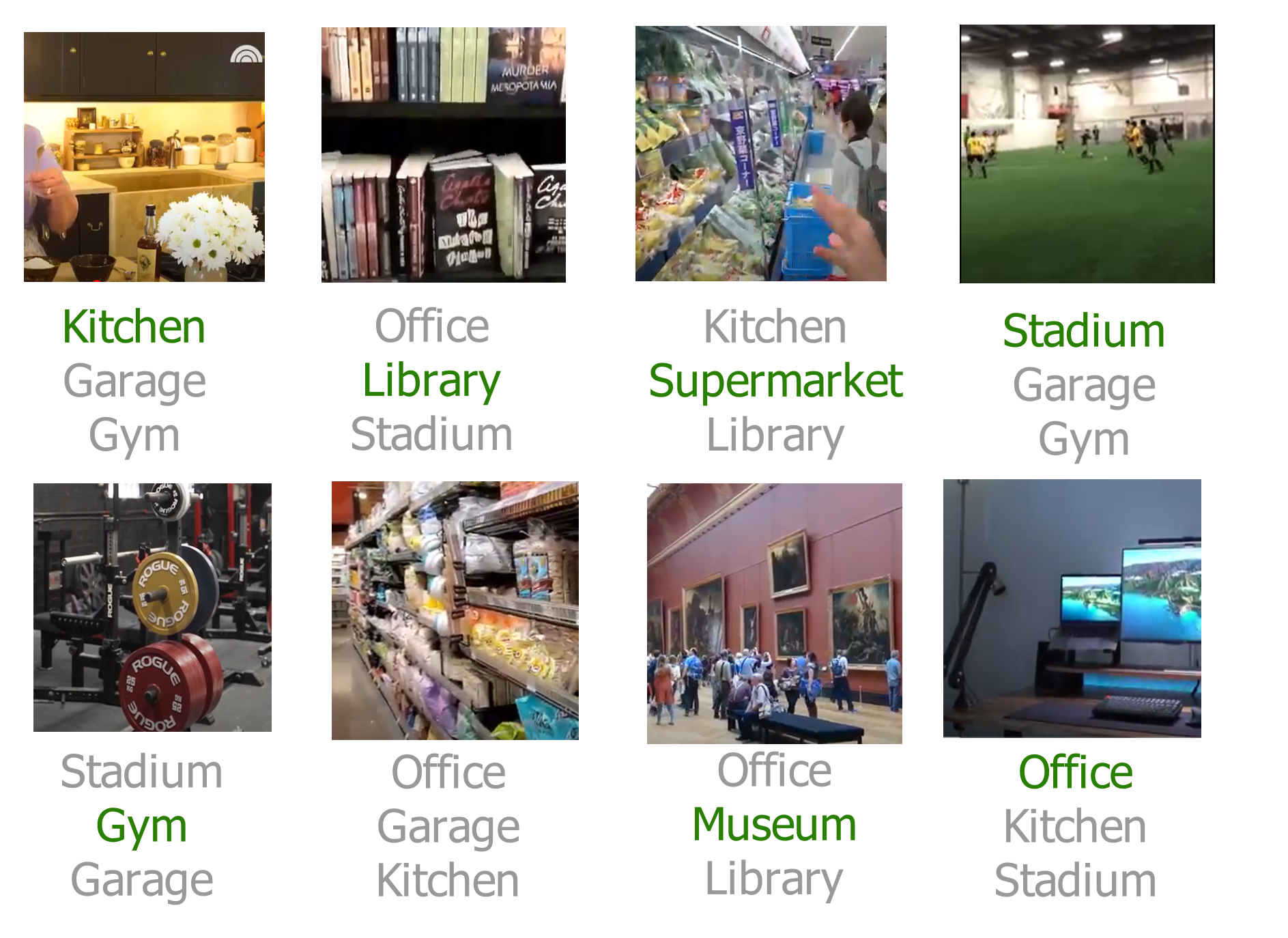}
\caption{Example of top predictions of the best performing model on the YouTubeIndoor subset. Frames sampled such that they are representative to the corresponding scene.}
\label{fig:yt_pred}
\end{figure}

Qualitatively, the confusion matrix in Figure \ref{fig:yt_cm} indicates misclassifications in classes such as `Office', `Gym', `Museum', and `Garage'. These spaces can vary significantly, e.g., due to personal decor or improvised home offices, or great variety in museum exhibits. Thus, it could be the case that these misclassifications occur due to misplaced objects which generally related to a different class, or multi-use spaces, such as small home gyms inside garages, etc. The classes with the lowest true positive percentages are `Office' and `Museum', with 44.3 and 52.7, respectively. 

When looking at specific videos, see Figure \ref{fig:yt_pred}, this idea is partly confirmed: the second image, a library, is confused for an office due to the books and papers present. Similarly, the third image, a supermarket, is assumed to be a kitchen due to the focus on produce. Within the top 2 predictions for the fourth image, a stadium, is the class garage, due to the variety of the predicted class. The multi-use of places such as garages and stadiums is once again highlighted, with them often being predicted for spaces related to sports or utility.

\section{Conclusions and Future Work}\label{sec:conclusion}

In this work, we contribute to the field of indoor scene recognition in videos by applying multi-modal deep learning techniques. On one side, we introduce InstaIndoor, a dataset composed of 3,788 videos collected from the Instagram platform that describe 9 different indoor scenes. Moreover, we define the YouTubeIndoor dataset, as a subset of the YouTube-8M dataset, composed of 900 videos describing 9 indoor scenes. Both datasets are suitable for visual and audio analysis of the depicted indoor scene in the respective videos.

We propose to assess the performance of several multi-modal architectures on these datasets. To this end, we perform an ablation study with respect to fusion strategies, as well as text and visual feature extraction techniques. Our proposed model achieves up to 70\% accuracy on our InstaIndoor dataset, a 59\% improvement with respect to random classification (11\%), and a minimum of 10\% improvement when compared to single modality approaches. On the YouTubeIndoor subset, our model achieves 74\% accuracy, which represents a 63\% improvement when compared to random classification (11\%). Our multi-modal approach obtains a significant performance boost with respect to single modality approaches. Single modality visual approaches perform relatively well using similar features to our top-performing model. On the other hand, single modality text approaches perform poorly likely due to a lack of sufficient information being transcribed. However, when joining the visual and text features, we can achieve a better overall performance, as the features complement one another. These results represent the baseline for these newly studied datasets, which we hope will encourage more works in this field. Moreover, they highlight not only the power of multi-modal learning for scene recognition, but also the potential for indoor scene recognition within the vast world of social media.

Overall, when comparing the results on the two datasets, we remark that certain models obtain poorer results on YouTubeIndoor than on InstaIndoor. This could be due to the video length, with YouTube videos being much longer, thus containing temporal dependencies and perhaps key explanations or objects which it may not be possible to capture within the number of frames with which we have experimented. A solution to alleviate this problem would be to crop the YouTube videos such that only the relevant scene is present, thus excluding segments from different scenes or heavily edited parts. Furthermore, we notice the relatively similar performance of Joint and Early fusion, often producing the best results, which can be attributed to factors such as final dense blocks and feature selection.


Our future lines of work include the extension of our newly proposed InstaIndoor dataset with new categories that capture other indoor scenes, aiming at scene recognition from the egocentric perspective of the mobile user. Moreover, it would also be interesting to experiment with different feature extraction techniques, aiming at capturing a more detailed description of the environment depicted in the video. We believe this will be welcome in fields such as forensics, marketing, or social sciences.

\begin{acknowledgements}
We would like to thank the Center for Information Technology of the University of Groningen for their support and for providing access to the Peregrine high performance computing cluster.
\end{acknowledgements}

\section*{Conflicts of Interest}
The authors declare that they have no conflicts of interest.

\bibliographystyle{splncs04.bst}
\bibliography{bibliography.bib}

\end{document}